\documentclass[conference]{IEEEtran}
\IEEEoverridecommandlockouts
\usepackage{cite}
\usepackage{amsmath,amssymb,amsfonts}
\usepackage{algorithmic}
\usepackage{graphicx}
\usepackage{textcomp}
\usepackage{xcolor}
\usepackage{booktabs}
\usepackage{fancyhdr}
\usepackage[colorlinks=true, urlcolor=blue, citecolor=black, linkcolor=black]{hyperref} 
\usepackage{comment}
\def\BibTeX{{\rm B\kern-.05em{\sc i\kern-.025em b}\kern-.08em
    T\kern-.1667em\lower.7ex\hbox{E}\kern-.125emX}}
\begin{document}

\title{Benchmarking Recurrent Event-Based Object Detection for Industrial Multi-Class Recognition on MTevent}

\author{
\IEEEauthorblockN{
Lokeshwaran Manohar,
Moritz Roidl
}

\IEEEauthorblockA{Chair of Material Handling and Warehousing, TU Dortmund University, Dortmund, Germany}

\IEEEauthorblockA{
lokeshwaran.manohar@tu-dortmund.de \\
moritz.roidl@tu-dortmund.de \\
}
}

\maketitle

\fancypagestyle{withfooter}{
\renewcommand{\headrulewidth}{0pt}
\fancyfoot[C]{\footnotesize Accepted to the Challenges and Opportunities of Neuromorphic Field Robotics and Automation IEEE ICRA Workshop - 2026}
}
\thispagestyle{withfooter}
\pagestyle{withfooter}

\begin{abstract}
Event cameras are attractive for industrial robotics because they provide high temporal resolution, high dynamic range, and reduced motion blur. However, most event-based object detection studies focus on outdoor driving scenarios or limited class settings. In this work, we benchmark recurrent ReYOLOv8s on MTevent for industrial multi-class recognition and use a non-recurrent YOLOv8s variant as a baseline to analyze the effect of temporal memory. On the MTevent validation split, the best scratch recurrent model (C21) reaches 0.285 mAP50, corresponding to a 9.6\% relative improvement over the non-recurrent YOLOv8s baseline (0.260). Event-domain pretraining has a stronger effect: GEN1-initialized fine-tuning yields the best overall result of 0.329 mAP50 at clip length 21, and unlike scratch training, GEN1-pretrained models improve consistently with clip length. PEDRo initialization drops to 0.251, indicating that mismatched source-domain pretraining can be less effective than training from scratch. Persistent failure modes are dominated by class imbalance and human-object interaction. Overall, we position this work as a focused benchmarking and analysis study of recurrent event-based detection in industrial environments.
\end{abstract}

\begin{IEEEkeywords}
Event camera, industrial robotics, object detection, recurrence, transfer learning, benchmarking
\end{IEEEkeywords}

\section{Introduction}
Event cameras report asynchronous brightness changes rather than dense image frames \cite{lichtsteiner2008firstevent, gallego2022survey}. Their low latency, high dynamic range, and reduced motion blur make them attractive for robotic perception under fast motion, challenging illumination, and partial occlusion. These properties are especially relevant for industrial automation, where sensors must often operate reliably despite motion, lighting variation, and scene clutter.

Despite this promise, event-based object detection has so far been evaluated mainly on a limited set of benchmark settings, with strong emphasis on autonomous driving datasets such as GEN1\cite{perot2020megapixelgen1} and related traffic-oriented scenarios. Other datasets, such as PEDRo \cite{boretti2023pedro}, broaden the application space toward robotics, but still focus on a single semantic category, namely person detection. As a result, comparatively little is known about how current event-based detectors behave in industrial multi-class environments with diverse object categories and strong class imbalance.

Industrial scenes differ from common driving benchmarks in several important ways. They require recognition of many object categories rather than only a few traffic participants, and they often contain cluttered layouts, repeated structures, partial occlusions, and frequent human-object interaction. In addition, rare classes may be severely underrepresented, making training and evaluation more sensitive to long-tail effects. These properties make industrial event-based detection a useful stress test for current model families.

They also motivate explicit study of temporal modeling. Because event evidence is sparse and distributed over time, single-window representations may provide only partial information about an object. Recurrent processing can in principle accumulate weak evidence across short temporal horizons and improve robustness under occlusion or fragmented motion cues. At the same time, longer temporal context is not guaranteed to help, since it may also introduce irrelevant history, noise accumulation, or temporal blur. It is therefore not obvious how much recurrence helps in industrial event-based detection, or which temporal horizon is most suitable.

In this paper, we study whether recurrent temporal modeling improves event-based object detection in industrial environments. We evaluate YOLOv8s\cite{ultralyticsyolo2023} and recurrent ReYOLOv8s \cite{silva2025reyolov8} variants on MTevent \cite{awasthi2025mtevent}, using 17 classes derived from the dataset annotations, and analyze three factors: (i) recurrence versus non-recurrence, (ii) temporal clip length, and (iii) event-domain pretraining and zero-shot transfer. Our contribution is not a new architecture or a new dataset. Instead, we adapt established model families and provide a focused benchmark and failure-case analysis for industrial multi-class recognition. This type of controlled evaluation is valuable because industrial event-based detection remains comparatively underexplored, even as benchmark diversity and application interest continue to expand. Code is available at \url{https://github.com/FLW-TUDO/ReYOLOv8-MTevent}.
\section{Related Work}

\subsection{Event-Based Detection and Representations}
Event-based object detection commonly converts asynchronous event streams into dense spatiotemporal tensors, such as voxel-grid style encodings, so that convolutional detectors can be reused \cite{smagulova2025survey}. Such representations preserve coarse temporal structure while remaining compatible with standard CNN backbones and detection heads. Recent progress in event-based detection has been driven largely by benchmark development and architectural adaptation in automotive scenarios \cite{smagulova2025survey}. Beyond recurrent CNN-based detectors, transformer-based event vision models such as RVT \cite{Gehrig_2023_rvt} have also reported strong performance in event-based perception tasks. However, these methods rely on different architectural assumptions and input processing pipelines, making direct controlled comparison outside the scope of this study. In this work, we adopt the same temporally binned event-tensor input format used in the ReYOLOv8 pipeline, allowing a controlled comparison between recurrent and non-recurrent variants without introducing a new event representation.

\subsection{Recurrent Temporal Modeling}
Temporal modeling is well motivated in event sensing because useful evidence is distributed over time rather than concentrated in a single frame. Recurrent modules such as ConvLSTM\cite{shi2015convlstm} can accumulate information across short temporal horizons while preserving spatial feature structure. ReYOLOv8 adopts this idea by augmenting a YOLOv8-style detector with recurrent processing for event-based object detection \cite{silva2025reyolov8}. Our goal is not to introduce a new recurrent architecture, but to evaluate how much recurrence helps in an industrial multi-class setting and how its effect compares with clip length and event-domain pretraining.

\subsection{Benchmarks and the Industrial Gap}
Prior event-camera benchmarks cover complementary settings, including driving-focused detection datasets such as GEN1, robotics-oriented person-detection datasets such as PEDRo, and indoor dynamic-scene datasets such as EVIMO2 \cite{boretti2023pedro,perot2020megapixelgen1,burner2022evimo2}. However, these settings do not fully match the industrial multi-class recognition problem considered here, where detectors must distinguish multiple object categories under clutter, repeated structures, partial occlusion, and frequent human-object interaction. This broader gap is also reflected in recent industrial robot perception benchmarks, which highlight realistic warehouse and factory challenges such as distant viewpoints, larger objects, varied scene configurations, and severe occlusion \cite{gouda2025mr6d}. MTevent provides a recent event-camera benchmark with dynamic object interactions, challenging viewpoints, occlusions, and 16 unique rigid objects together with moving-object annotations \cite{awasthi2025mtevent}. These properties make it a useful testbed for studying temporal context, transferability, and long-tail failure modes in industrial event-based recognition.
\section{Method}

\subsection{Task Setup and Input Encoding}
We perform multi-class object detection on MTevent using 5-channel temporally binned event-tensor inputs (5 temporal bins, 50\,ms windows, 256$\times$320). Models are trained and evaluated with consistent preprocessing across all ablations.

\subsection{Model Family}
We use the standard YOLOv8s architecture as the common base model, adapted to 5-channel event input. The YOLOv8s variant is chosen to balance computational cost and detection performance, enabling controlled comparison across recurrent and non-recurrent settings within a fixed training budget.

\textbf{YOLOv8s baseline.} Non-recurrent detector operating on single temporal windows (clip length 1).

\textbf{ReYOLOv8s.} Recurrent detector that augments the YOLOv8s backbone with ConvLSTM modules at intermediate feature stages, enabling temporal aggregation of event features across clip sequences. We evaluate clip lengths $\{3,7,11,21\}$ with stride-based temporal sampling, where clip elements are selected from temporally ordered event windows at a fixed interval rather than from consecutive windows. Hidden states are propagated within each sampled clip sequence and reset between independent training or evaluation samples.

This work uses adapted existing architectures and does not claim a novel detector design.

\subsection{Pretraining and Transfer Settings}
We evaluate recurrent initialization from two external event datasets:
\begin{itemize}
    \item \textbf{GEN1 init}: driving-oriented event detection pretraining with car and pedestrian categories.
    \item \textbf{PEDRo init}: robotics-oriented event dataset for person detection.
\end{itemize}

\subsection{Training Details}
All experiments use AdamW, $\texttt{lr0}=0.00181$, cosine LR decay, batch size 16, image size 320, and early-stopping patience 50. Model selection and early stopping are based on validation performance. Evaluation reports mAP50 and mAP50--95 on the validation split.

To keep comparisons controlled, we keep detector head configuration and post-processing consistent across all runs. NMS and confidence thresholds are unchanged between recurrent and non-recurrent settings. This protocol isolates temporal modeling and initialization effects from confounding hyperparameter changes. GEN1-pretrained variants are evaluated across all clip lengths $\{3,7,11,21\}$ to study the interaction between pretraining and temporal context. PEDRo-pretrained variants are evaluated at clip length 11 and 21.
\section{Experiments and Results}

\subsection{Dataset: MTevent}
MTevent is a recent industrial event-camera benchmark recorded in warehouse and factory-like scenes using a stereo event camera together with an RGB camera \cite{awasthi2025mtevent}. The dataset contains 75 scenes in total, and we use 17 detection classes comprising 16 rigid object categories plus \textit{human}. For our experiments, we restrict the data to the left event stream and use a custom scene-level train/validation/test split of 60/13/2 scenes, which was fixed prior to model development to avoid data leakage or split tuning, as MTevent does not define a standardized evaluation split. Due to the small size and limited class coverage of the test partition, we report validation performance as the primary benchmark and use test-set results of the best model only as a reference point. The dataset is strongly class-imbalanced, making it a challenging benchmark for studying recurrence, transferability, and long-tail failure modes in industrial event-based detection.

\subsection{Main Ablation: Recurrence, Clip Length, and Pretraining}
\begin{figure}[t]
\centering
\includegraphics[width=0.92\columnwidth]{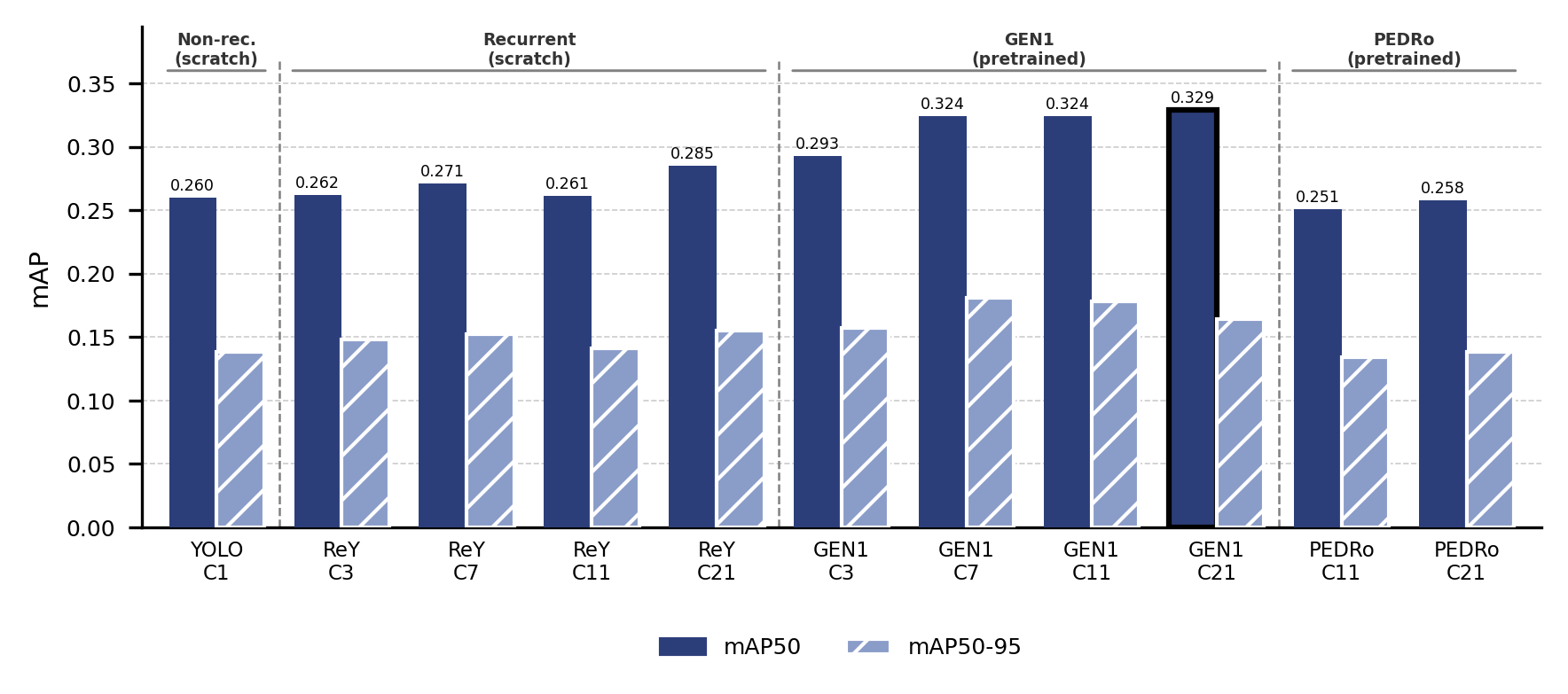}
\caption{Validation performance on MTevent across non-recurrent and recurrent settings, temporal clip lengths, and pretraining configurations. YOLO denotes the non-recurrent YOLOv8s baseline, ReY-C$n$ denotes ReYOLOv8s trained from scratch with clip length $n$, and GEN1/PEDRo-C$n$ denote ReYOLOv8s fine-tuned from pretrained weights at clip length $n$. GEN1 initialization yields the best overall result at C21. Numerical values are reported in Table~\ref{tab:ablation}.}
\label{fig:ablation}
\end{figure}

Table~\ref{tab:ablation} and Fig.~\ref{fig:ablation} summarize the validation benchmark. The non-recurrent YOLOv8s baseline establishes the reference performance for single-window event detection. ReYOLOv8s models trained from scratch quantify the effect of recurrent temporal modeling across clip lengths, while pretrained recurrent variants evaluate transfer from external event domains.

Across scratch-trained models, recurrence improves over the non-recurrent baseline, with the best scratch recurrent setting reaching a 9.6\% relative mAP50 improvement. The best scratch recurrent setting, clip length 21, reaches 0.285 mAP50 and 0.155 mAP50--95, compared with 0.260 and 0.138 for the non-recurrent YOLOv8s baseline. This indicates that temporal context is beneficial for industrial event-based detection, but recurrence alone is not sufficient to close the remaining performance gap.

Performance does not increase monotonically with clip length, as C11 slightly underperforms C7 before performance improves again at C21. One possible explanation is that intermediate sequence lengths may already increase temporal optimization difficulty while still not covering a sufficiently complete motion pattern to provide stable additional cues. In contrast, longer sequences such as C21 may better capture object motion and interaction phases. 

Pretraining has a strong but source-dependent effect on MTevent performance. Fine-tuning from GEN1 yields the best overall result of 0.329 mAP50 at clip length 21, substantially outperforming all scratch-trained variants. Unlike the non-monotonic clip-length behaviour observed for scratch-trained models, GEN1-pretrained models improve consistently with clip length (C3: 0.293, C7: 0.324, C11: 0.324, C21: 0.329), suggesting that pretraining stabilizes temporal optimization and enables the model to exploit longer temporal context. In contrast, PEDRo-pretrained models reach only 0.251 mAP50 at clip 11 and 0.258 at clip 21 — below both the best scratch-trained recurrent model and the non-recurrent YOLOv8s baseline (0.260), and showing negligible improvement with longer clips. This indicates that event-domain pretraining is not universally beneficial: when the source dataset is mismatched in scene structure, object categories, or labeling density, transfer can hurt rather than help. The results suggest that source-domain alignment matters more than modality similarity alone.

All reported results are based on single training runs; therefore, small differences below approximately 0.01 mAP50 should be interpreted with caution, as they may fall within normal training variance.

\begin{table*}[t]
\caption{ABLATION RESULTS ON THE MTEVENT VALIDATION SPLIT COMPARING NON-RECURRENT AND RECURRENT MODELS, TEMPORAL CLIP LENGTH, AND EVENT-DOMAIN PRETRAINING.}
\label{tab:ablation}
\centering
\begin{tabular}{lccccc}
\toprule
Model & Recurrent & Init & Clip & mAP50 & mAP50-95 \\
\midrule
YOLOv8s baseline & No & Scratch & 1 & 0.260 & 0.138 \\
ReYOLOv8s & Yes & Scratch & 3 & 0.262 & 0.148 \\
ReYOLOv8s & Yes & Scratch & 7 & 0.271 & 0.152 \\
ReYOLOv8s & Yes & Scratch & 11 & 0.261 & 0.141 \\
ReYOLOv8s & Yes & Scratch & 21 & 0.285 & 0.155 \\
\midrule
ReYOLOv8s (GEN1 init) & Yes & GEN1 & 3 & 0.293 & 0.157 \\
ReYOLOv8s (GEN1 init) & Yes & GEN1 & 7 & 0.324 & 0.181 \\
ReYOLOv8s (GEN1 init) & Yes & GEN1 & 11 & 0.324 & 0.178 \\
ReYOLOv8s (GEN1 init) & Yes & GEN1 & 21 & \textbf{0.329} & 0.164 \\
ReYOLOv8s (PEDRo init) & Yes & PEDRo & 11 & 0.251 & 0.134 \\
ReYOLOv8s (PEDRo init) & Yes & PEDRo & 21 & 0.258 & 0.138 \\
\bottomrule
\end{tabular}
\end{table*}

\subsection{Per-Class Trends}
Strong classes include \textit{wooden\_pallet}, \textit{human}, \textit{ikea\_skubb\_case}, and \textit{ikea\_vesken\_trolley}, whereas weak classes include \textit{amazon\_luggage}, \textit{ikea\_moppe\_drawer}, \textit{ikea\_labbsal\_basket}, and \textit{small\_klt}. These per-class trends should be interpreted cautiously as descriptive tendencies rather than stable rankings, given the strong class imbalance of MTevent and the fact that all results are based on single training runs. They are broadly consistent with class imbalance, limited training diversity, and low event-space separability for visually similar or weakly structured objects.

The observed class differences also plausibly align with object geometry and motion signature. Large rigid items with clear edges may generate more stable event contours and support more consistent localization across temporal bins. By contrast, small containers and visually similar storage objects can produce sparse or overlapping activations, increasing confusion between nearby classes.

\begin{figure}[t]
    \centering
    \includegraphics[width=0.32\linewidth]{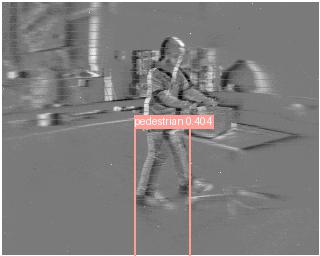}\hfill
    \includegraphics[width=0.32\linewidth]{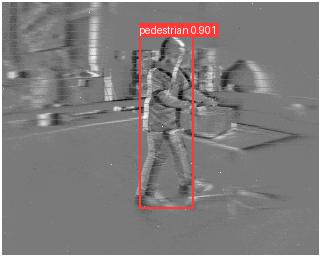}\hfill
    \includegraphics[width=0.32\linewidth]{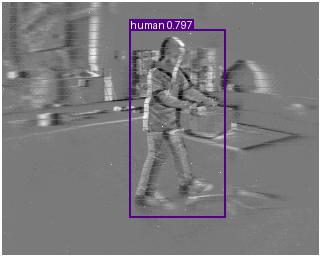}
    \caption{Qualitative zero-shot transfer examples on MTevent. The GEN1-pretrained model does not transfer reliably to MTevent humans, whereas the PEDRo-pretrained model produces a stronger human detection. The fine-tuned MTevent model detects the human with the correct target label.}
    \label{fig:intro}
\end{figure}
\subsection{Zero-Shot Transfer}
We additionally evaluate zero-shot transfer from GEN1- and PEDRo-pretrained recurrent models to MTevent using the shared \textit{human} category only. Qualitative examples are shown in Fig.~\ref{fig:intro}. We emphasize that this analysis is not intended to assess recognition of unseen classes, but rather to examine transfer behavior for a shared semantic category under domain shift.

Before MTevent fine-tuning, zero-shot transfer remains limited. The GEN1-pretrained model does not produce reliable human detections on MTevent (AP50 = 0.000), while the PEDRo-pretrained model achieves AP50 = 0.136 for the human class, with consistent full-body detections at confidences between 0.82 and 0.91. This confirms that source-domain alignment with robotics-oriented person data enables limited zero-shot generalisation, while driving-oriented pretraining does not transfer to the human category in industrial scenes. This difference suggests that zero-shot transfer depends strongly on source-target similarity even for shared semantic categories, rather than on label overlap alone.

We treat the explanation for this behavior as a hypothesis rather than a confirmed causal finding. A likely factor is mismatch between source and target data characteristics, including scale, motion pattern, viewpoint, and human-object interaction. In particular, workers carrying objects in MTevent can produce merged or distorted event signatures that differ from cleaner person-only detections in the source domain. Overall, these zero-shot observations are consistent with the quantitative fine-tuning results, which indicate that source-domain similarity plays a major role in transfer effectiveness.

\subsection{Industrial Failure Cases}
Two failure patterns remain dominant. First, rare classes remain difficult due to long-tail class scarcity and limited training diversity. Second, human-object carrying scenarios often produce merged event blobs or overlapping activations, which reduces class separability and makes co-occurring detections more difficult.

These failure cases are especially relevant for deployment because they arise in scenarios that matter most in industrial environments: worker-assisted manipulation, mixed human-object motion, and cluttered activity near robotic systems. We therefore interpret them not only as benchmark errors, but as practically important limitations of the current detection pipeline.
\section{Conclusion}
We presented a compact benchmark of recurrent event-based object detection for industrial multi-class recognition on MTevent. Relative to the non-recurrent YOLOv8s baseline, the best scratch recurrent setting improves mAP50 from 0.260 to 0.285 (9.6\% relative gain), while the best overall result of 0.329 mAP50 is achieved by GEN1-pretrained ReYOLOv8s at clip length 21. Temporal clip length affects scratch-trained performance in a non-monotonic manner, whereas GEN1-pretrained models improve consistently with clip length, suggesting pretraining stabilizes temporal optimization. Event-domain pretraining has a stronger impact than recurrence alone, though its benefit depends critically on source-domain alignment. Transfer analysis shows that even for shared categories (e.g., human), zero-shot performance depends strongly on how similar the source and target data are, rather than on label overlap alone.

Overall, the benchmark clarifies practical strengths and bottlenecks of recurrent event-based detection in industrial settings. In particular, the results suggest that, in the current setting, improved data coverage and better transfer protocols may offer more immediate gains than additional architectural complexity. At the same time, the study remains limited by strong class imbalance in MTevent, the use of only the left event stream with a custom split, the absence of broader event-based baseline comparisons, and the small number of validation and test scenes.

From a workshop perspective, we hope these results provide a useful and reproducible reference point for future industrial event-camera studies. Immediate next steps include broader baseline comparisons, including transformer-based event detectors\cite{Gehrig_2023_rvt}, and more controlled cross-domain transfer experiments. Extending the benchmark with improved class coverage, additional sensor streams, and possibly synthetic data augmentation are promising directions for future work.

\bibliographystyle{IEEEtran}
\bibliography{references}

\end{document}